# Estimating Genome Reversal Distance by Genetic Algorithm

**Andy Auyeung**
Oklahoma State University
Math Science 219
Stillwater, OK 74078
1 (405) 744–5668
wingha@cs.okstate.edu

**Ajith Abraham**
Oklahoma State University
North Hall 328
Tulsa, OK 74106
1 (918) 594–8188
aa@cs.okstate.edu

**Abstract-** Sorting by reversals is an important problem in inferring the evolutionary relationship between two genomes. The problem of sorting unsigned permutation has been proven to be NP-hard. The best guaranteed error bounded is the 3/2-approximation algorithm. However, the problem of sorting signed permutation can be solved easily. Fast algorithms have been developed both for finding the sorting sequence and finding the reversal distance of signed permutation. In this paper, we present a way to view the problem of sorting unsigned permutation as signed permutation. And the problem can then be seen as searching an optimal signed permutation in all $2^n$ corresponding signed permutations. We use genetic algorithm to conduct the search. Our experimental result shows that the proposed method outperform the 3/2-approximation algorithm.

## 1 Introduction

Genome Rearrangement is a mechanism that happens in mitochondrial genomes (Russell 2002). The genes order in mitochondrial genome is constantly under rearrangement. Therefore, by estimating the rearrangement distance between two genomes, the relationship between them can also be estimated (Pevzner 2001). Reversal is the most commonly seen mechanism that genomes are rearranged. Figure 1 shows the estimated transformation from *Tobacco* to *Lobelia fervens* by reversals (Bafna and Pevzner; 1996). There are two variations of this problem, signed permutation and unsigned permutation. For unsigned permutation, a genome is modeled as a permutation $\pi$ with order $n$ (i.e. a permutation of {1, 2, …, n}), where $n$ is the number of gene blocks in the genome. Let the permutation $\pi = \pi[1]$ $\pi[2]$ … $\pi[n]$, the reversal operation $\rho(i,j)$ rearrange $\pi$ into $\pi[1]$ … $\pi[i-1]$ $\pi[j-1]$ … $\pi[i]$ $\pi[j]$ … $\pi[n]$. For signed permutation $\pi'$, each $\pi'[k]$ has either a positive or a negative sign. Each reversal operation $\rho(i,j)$ not only rearrange $\pi'$ but also negate the sign of $\pi'[k]$ for $i \le k < j$. The problem of estimating reversal distance between two genomes is formulated as sorting permutation by reversal operation. That is, given $\pi$ (or $\pi'$), we want to find a sorting sequence that uses minimum number of reversal to sort $\pi$ (or $\pi'$) into identity permutation (i.e. the permutation, 1 2 … $n$ for unsigned permutation, and +1 +2 … +$n$ for signed permutation). We called the minimum number of reversal the reversal distance.

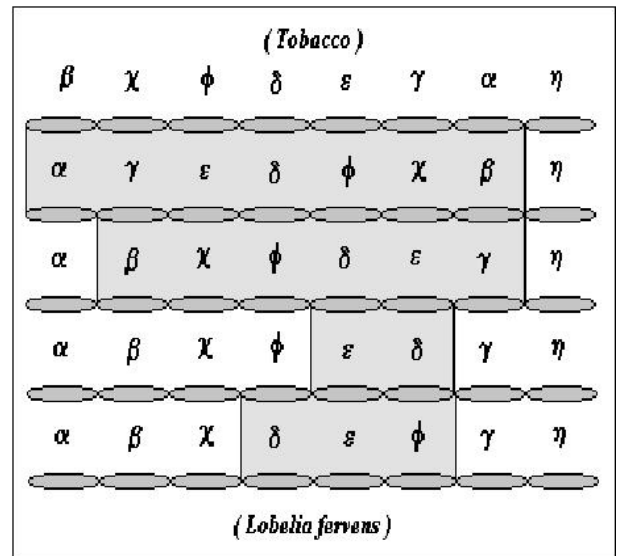

Figure 1. Transformation from *Tabacco* to *Lobelia fervens* by reversals (Bafna and Pevzner; 1996).

The problem of sorting signed permutation can be solved in $O(n^2)$ time (Kaplan et al. 1997). The problem of finding the reversal distance of signed permutation can be solved in $O(n)$ time (Bader et al. 2001). However, both sorting and finding the reversal distance of unsigned permutation has been proven to be NP-hard (Caprara 1997). So, error bounded heuristic solutions have been proposed (Bafna and Pevzner 1996; Kececioglu and Sankoff 1995). The lowest guaranteed error bound thus far is the 3/2-approximation algorithm (Christie 1998). The 3/2-approximation algorithm uses the fact that any cycle decomposition of the breakpoint graph that maximize the number of 2-cycles exists a sorting sequence with length at most 3/2 of the optimal sorting sequence.

In this paper, we propose a genetic algorithm for sorting unsigned permutation by reversal. Our method does not provide guaranteed error bound. However, our experiment shows that it finds better solution than the 3/2-

approximation algorithm. Also, so far, all heuristic algorithms for this problem use a constructive manner to find the solution. We would like to show an alternative approach how this problem can be solved in inductive manner.

The rest of the paper is organized as the following. In Section 2, some background materials on sorting permutation by reversal and the concept of genetic algorithm are presented. In Section 3, the proposed method is explained. In Section 4, the experimental setup and results are shown. In Section 5, observations from the experiment are discussed. Finally in Section 6, some concluding remarks are made.

## 2 Reviews

### 2.1 Breakpoint Graph

An unsigned permutation can be modeled by a breakpoint graph. For each gene (a number in the permutation), we will create a node for it. The idea of breakpoint graph is to mark the desired and realistic relationship between these nodes in the permutation. For each pair of the nodes we draw a black edge between them if they are adjacent in the permutation, and we draw a red edge between them if they are adjacent in the identity permutation. In order to model the orientation, we expand the unsigned permutation to have a zero at the front and a $n+1$ at the end. An example is shown in Figure 2. It has been shown that given a cycle decomposition of the breakpoint graph, any reversal can at most change the number of cycles by one. Besides, it has also been shown that given a cycle decomposition of the breakpoint graph, the corresponding shortest sorting sequence can then be found. Thus, the key problem is to find a cycle decomposition that provides the shortest sorting sequence for the unsigned permutation. However, the problem of finding an optimal cycle decomposition is NP-hard.

On the other hand, sorting signed permutation can be solved easily. We first create the breakpoint graph as above. Then for each gene node, we split it into two nodes according to its sign (and we shift to number of the node accordingly). An example is shown in Figure 3. The advantage is that now each node has exactly one red edge and one black edge associated with it. Thus, there is only one cycle decomposition of this breakpoint graph. Many algorithms have been proposed to find the sorting sequence. The best known time bound is $O(n^2)$ time. And the best known time bound time for finding the reversal distance is $O(n)$.

### 2.2 Genetic Algorithm

Genetic algorithm is a searching technique. The idea is inspired by natural selection happens in evolution. A genetic algorithm works with a population of individuals, each representing a solution to a given problem. The idea is to use evolutionary model to determine the next searching area from that is the next population, is generated by selection, crossover and mutation operation according to the fitness of solutions in the current population.

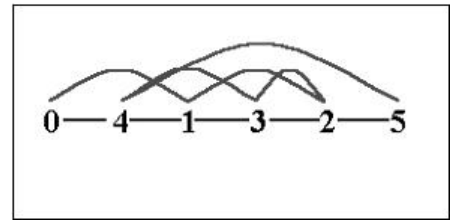

Figure 2. Breakpoint graph for unsigned permutation 4 1 3 2.

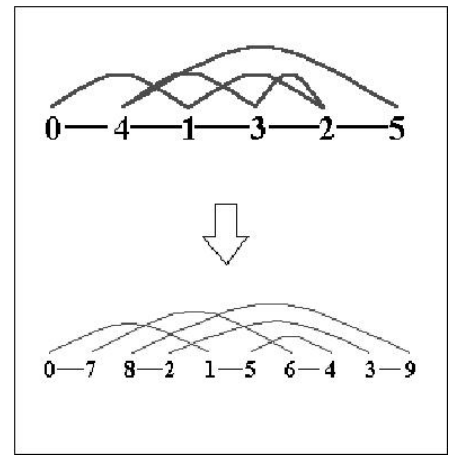

Figure 3. Breakpoint graph for signed permutation +4 -1 +3 -2.

The norm of the genetic algorithm can be describe as the following. A population of possible solutions is initially generated. The algorithm is divided in generations. In each generation, if the termination condition has not met, next population will be first determined by selection and crossover. In selection, individuals in the current population are probabilistically selected according to their fitness to move to the next population. In crossover, pairs of individuals are probabilistically selected according to their fitness to generate a new pair of individuals (offspring) by the crossover operator. Once the next population has been determined, then mutation operation will be probabilistically applied to individuals in the next population. Finally, we re-label the next population to be the current population that is to symbolize the old population has died out. And the fitness of the new individuals will be evaluated. In each generation, only fit individuals can produce offspring and survive. Thus, a population of fit solutions would be expected when the algorithm terminates. And the best individual would be used as the solution to the problem.

## 3 Proposed Method

The idea of the proposed method is to view all the possible cycle decomposition of the unsigned permutation as the signed permutations that have the same gene order. Except node 0 and $n+1$, every node in the breakpoint graph has degree 2 for red edges and also for black edges. So, a cycle decomposition is used to define the (color) alternating paths. However, there is a corresponding signed permutation that actually defines the same alternating paths. Therefore, we can now turn our focus on signed permutation instead of cycle decomposition. Define the set *Signed($\pi$)* be the set of signed permutations that have the same gene order as $\pi$. For example, when $\pi$ is 2 1, then *Signed($\pi$)* is { -2 -1, -2 +1, +2 −1, +2 +1}. Thus, the size of *Signed($\pi$)* is $2^n$. The following two observations are required for our method.

> *Observation 1:*
> *Each $\pi' \in Signed(\pi)$ can deduces a valid sorting sequence for $\pi$.*
>
> *Proof:*
> *Let sorting sequence $\rho$ sort $\pi'$ into identity (i.e. $\pi'_\rho = id$). Because $|\pi'[i]| = \pi[i]$, then $\pi'_\rho[i] = \pi_\rho[i]$, for all i. Thus, $\rho$ can also sort $\pi$ (i.e. $\pi_\rho = id$).*

> *Observation 2:*
> *There exist $\pi^* \in Signed(\pi)$ that deduces an optimal sorting sequence for $\pi$.*
>
> *Proof:*
> *Let $\rho$ be a sorting sequence for $\pi$ that uses minimum number of reversals. For each $\pi[k]$, let count[k] be the number of times that $\pi[k]$ is included in reversals of $\rho$, i.e. $\rho(i,j)$ $i \leq k < j$. Then $\pi^*$ is the following, $\pi^*[k]$ has a positive sign if count[k] is even, otherwise it has a negative sign. Because $|\pi'[i]| = \pi[i]$, $\pi_\rho[i] = |\pi'_\rho[i]|$, for all i. However, all $\pi'_\rho[i]$, must be positive by our construct. Thus $\rho$ can also $\pi^*$ and the reversal distance for $\pi^*$ is equal to the reversal distance for $\pi$.*

From the two observations, we can see that the problem can be solved in O($2^n$ n) time. That is to find the reversal distance for all $2^n$ $\pi'$. Let $\pi^*$ be the $\pi'$ that has minimum reversal distance. Then the sorting sequence of $\pi$ is the sorting sequence of $\pi^*$. However, it is not feasible to go through all $2^n$ $\pi'$. Thus, we use genetic algorithm to find $\pi^*$. There is no guarantee that the genetic algorithm would find $\pi^*$, however, we could expect the genetic algorithm would find a $\pi'$ that has low reversal distance.

## 4 Experiment

The genetic algorithm is the following. We allow the population size to be $n^2$. The initial population is randomly generated binary strings representing the signs of the genes in the permutation. However, we apply a heuristic on taking all trivial cycles (i.e. cycles that compose of exactly one red edge and one black edge). Thus, sorted substrings would be assigned to the same sign (positive for ascending sorted substring; negative for descending sorted substring). The fitness is evaluated by the reversal distance of this signed permutation. Single point crossover and mutation are used with rate 0.5 and 0.3 respectively. And the genetic algorithm is terminated when the best reversal distance in the population remains unchanged in three generations.

We conduct the experiment by randomly generated permutations, where permutations are generated by performing n random swap operations on the identity permutation. Figure 4 shows the comparison between the 3/2-approximation algorithm and the proposed method. (The actual data is shown in Appendix A.) The figure shows the number of reversals required in the sorting sequences found by the two methods. The comparison is on the average solution of ten runs with *n* is between 10 to 150. We can see that the proposed method produce better solution than the 3/2-approximation algorithm.

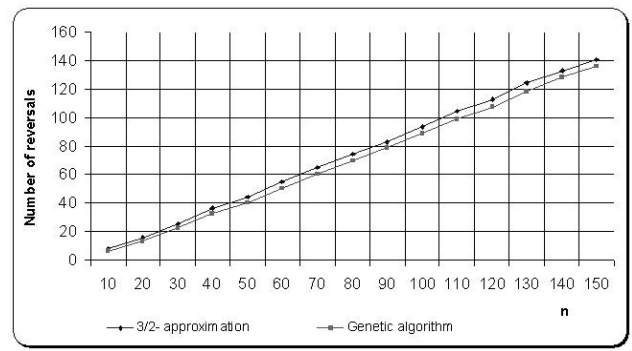

Figure 4. Performance comparison of 3/2-approximation and genetic algorithm.

## 5 Discussion

Sorting unsigned permutation has a trivial *n*-1 upper bound. That is, simply using one reversal to put one gene block in place. When *n*-1 gene blocks are in place, the *n*- th block is already in place. In fact, this is the reversal distance for the "hard-to-sort" Gollan permutation $\gamma_n$ (and $\gamma_n^{-1}$). From Figure 4, we can see that the solutions found by both the proposed method and the 3/2-approximation are not very far from the $n$-1 upper bound. That observation is consistent with the analysis done by Bafna and Pevzner that says the reversal distance between 2 random permutations is very close to the $n$-1 upper bound.

Figure 4 shows that the solutions produced by both the proposed method and the 3/2-approximation are growing linearly with $n$. Although the experiment shows that the proposed method only outperform the 3/2-approximation by a small distance, we can see that the 3/2-approximation has a steeper slope than the proposed method. In another word, the improvement from the proposed method scales up with $n$.

During our implementation, we found our heuristically initialized population enhances the average performance of the genetic algorithm. In fact, sorted substrings do not always stay together. Notice when we optimally sort the permutation 3 4 1 2, one of the sorted substrings (i.e. 3 4 or 1 2) has to be broken. However interestingly, it has been proven that there exists a sorting sequence that does not break sorted substring with length longer than 2 (Hannenhalli and Pevzner, 1996).

Our genetic operators and parameters to the genetic algorithm are not carefully chosen. The main goal of this paper is to introduce the representation that the sorting unsigned permutation problem can be solved by genetic algorithm. When selecting the genetic operators and parameters of the genetic algorithm, we believe it should be domain specific. For example, a mathematician may be interested in purely random permutations; while a biologist may be interested in biological shuffled permutations. And it is natural to expect that different permutation groups have different characteristics in their breakpoint graphs. So, different heuristic would be required in the genetic algorithm.

The computational time was not our major concern during the implementation of the proposed method. So, we used the $O(n^2)$ time Java implementation created by Itsik Mantin to find the reversal distance of signed permutations for the fitness evaluation (Mantin and Shamir, 1999). However, $O(n)$ time algorithm is known and can be used. Here we provide a time complexity comparison. The 3/2-approximation requires $O(n^2)$ time. On the other hand, the proposed method requires $O(n^2)$ size of population; each requires $O(n)$ time to evaluate its fitness in each generation; and there can be at most $O(n)$ generations (because the upper bound for reversal distance is $n$-1 and the genetic algorithm enforces improvement on the best reversal distance in every 3 generations otherwise terminates). Therefore, the proposed method requires $O(n^4)$ time.

# 6 Conclusion

Sorting unsigned permutation by reversals served an important role in inferring evolutionary history. A 3/2-approximation has been proposed and is the lowest guaranteed error bound thus far. All previous methods use constructive approach to produce the solution. This paper introduces a new inductive approach that uses genetic algorithm to find the solution. Experimental result shows that the proposed method outperforms the 3/2-approximation algorithm, although it does not mathematically guarantee the quality of the solution. Due to the inductive perspective, different searching methods can then be applied to solve the problem. Also, this perspective creates possibilities to estimate the reversal distance by sampling the solution space. Such estimation allows the estimated answer on both higher and lower sides, while the previous methods always produce over estimated answer.

# Acknowledgements

We would like to thank David Christie for all his help during our implementation of the 3/2-approximation algorithm. We also would like to thank Itsik Mantin and Ron Shamir make available to public the Java applet and the source code for the sorting signed permutation algorithm.

## Appendix A

| $n$ | 3/2-approximation | Genetic algorithm |
|---|---|---|
| 10 | 7.6 | 5.9 |
| 20 | 15.8 | 13.3 |
| 30 | 25.8 | 22.2 |
| 40 | 36.1 | 32.1 |
| 50 | 43.9 | 40.4 |
| 60 | 54.9 | 50.6 |
| 70 | 65.1 | 60.4 |
| 80 | 73.9 | 69.3 |
| 90 | 82.8 | 78.7 |
| 100 | 93.5 | 89.2 |
| 110 | 104.5 | 99.0 |
| 120 | 112.8 | 107.8 |
| 130 | 124.1 | 118.0 |
| 140 | 132.9 | 128.1 |
| 150 | 140.7 | 136.0 |

Table 1. Permformance comparison of 3/2-approximation and genetic algorithm.